\documentclass{article}

\usepackage{PRIMEarxiv}

\usepackage[utf8]{inputenc} 
\usepackage[T1]{fontenc}    
\usepackage{hyperref}       
\usepackage{url}            
\usepackage{booktabs}       
\usepackage{amsfonts}       
\usepackage{nicefrac}       
\usepackage{microtype}      
\usepackage{lipsum}
\usepackage{fancyhdr}       
\usepackage{graphicx}       
\graphicspath{{media/}}     
\usepackage{algorithm}
\usepackage{algorithm}
\usepackage{algpseudocode}
\usepackage{amsmath}

\title{I2I - STRADA – Information to Insights via Structured Reasoning Agent for Data Analysis}

\author{
  Sai Barath Sundar, Pranav Satheesan, Udayaadithya Avadhanam \\
  Mphasis Limited \\
  \texttt{\{sai.sundar, pranav.satheesan, udayaadithya.a\}@mphasis.com} \\
}

\begin{document}
\maketitle

\begin{abstract}
Recent advances in agentic systems for data analysis have emphasized automation of insight generation through multi-agent frameworks, and orchestration layers. While these systems effectively manage tasks like query translation, data transformation, and visualization, they often overlook the structured reasoning process underlying analytical thinking. Reasoning large language models (LLMs) used for multi-step problem solving are trained as general-purpose problem solvers. As a result, their reasoning or thinking steps do not adhere to fixed processes for specific tasks. Real-world data analysis requires a consistent cognitive workflow: interpreting vague goals, grounding them in contextual knowledge, constructing abstract plans, and adapting execution based on intermediate outcomes. We introduce I2I-STRADA (Information-to-Insight via Structured Reasoning Agent for Data Analysis), an agentic architecture designed to formalize this reasoning process. I2I-STRADA focuses on modeling how analysis unfolds via modular sub-tasks that reflect the cognitive steps of analytical reasoning. Evaluations on the DABstep and DABench benchmarks show that I2I-STRADA outperforms prior systems in planning coherence and insight alignment, highlighting the importance of structured cognitive workflows in agent design for data analysis.
\end{abstract}

\section{Introduction}
\label{sec:intro}
Real-time and ad-hoc data analysis in enterprise environments is a complex task as data tends to be heterogeneous, non-standard and lacking quality. This is due to the diversity of systems, the variability of human input, and the continuous evolution of business processes \cite{rozony2024systematic}. Traditionally, various data harmonization techniques have been used to address challenges arising from data in multiple formats, incompleteness, and missing values \cite{Cheng2024-uc}. While dealing with multiple sources, the same entity can have conflicting attributes due to naming conventions, out-of-date data etc., necessitating a truth discovery process before proceeding with any further analysis \cite{li2015surveytruthdiscovery}. Furthermore, as organizational processes expand, changes to the data structures and corresponding analytical requirements result in significant re-engineering efforts \cite{PUTRAMA2024110853}, \cite{Bandara2022-vu}. Thus, it is imperative that data analytics systems incorporate procedural knowledge and are knowledge-driven \cite{Bandara2022-vu}. 

LLMs are naturally suited to address these challenges, given their ability to understand unstructured data, infer context, and adapt to evolving semantics across heterogeneous sources. In \cite{santos2025interactivedataharmonizationllm}, the authors focus on developing a system for data harmonization of tabular data sources using LLMs. In \cite{chen2023symphony}, the authors leverage representation learning techniques for multi-modal data discovery and subsequently query decomposition for planning and execution. In other similar works like \cite{wang2025operationalizingheterogeneousdatadiscovery}, \cite{wang2025aop} the authors formalize a set of multi-modal semantic operators which are composed into execution pipelines. While these LLM based methods focus on tasks like query translation or data transformation, they fail to combine them into a structured reasoning process required for analytical thinking. 

We introduce \textbf{I2I – STRADA} that enables going from \textbf{I}nformation to \textbf{I}nsights via a \textbf{St}ructured \textbf{R}easoning \textbf{A}gent for \textbf{D}ata \textbf{A}nalysis. The agent follows a workflow composed of multiple specialized sub-tasks, each responsible for a distinct aspect of reasoning and planning. We discuss related work and key limitations in the next section followed by the details of our approach. We evaluate I2I-STRADA on DABstep \cite{egg2025dabstepdataagentbenchmark} and DABench \cite{hu2024infiagentdabenchevaluatingagentsdata} data analysis benchmarks that focus on scenarios where agents must operate under procedural constraints and deliver insights. Results show significant improvements in planning quality and alignment with analytical objectives, underscoring the value of structured reasoning in agent design. 

\section{Related work}
\label{sec:relatedwork}
Recent contributions to data analysis agents can be categorized into two main streams: (1) those focused on planning, and (2) those aimed at building agents for end-to-end analytics platforms.

\subsection{Planning focused approaches}
DatawiseAgent \cite{you2025datawiseagentnotebookcentricllmagent}, employs a (Depth First Search) DFS like planning and incremental code execution mechanism along with self-debugging capabilities. This approach is proposed to address the complexities involved in solution exploration and ensuring the result of code execution is consistent with the corresponding planning step. However, the lack of global planning can result in inconsistencies in the trajectories generated on the fly. DataInterpreter \cite{hong2024datainterpreterllmagent}, aims to produce global execution steps by generating a graph of tasks for a given problem. The tasks are chosen from a list of fine-grained task definitions most seen in data processing and data science pipelines. However, both the methods above do not incorporate a data understanding step, thereby increasing the chances of erroneous interpretation of data elements and domain-specific computations. 

\subsection{Agents for end-to-end analytics platforms}
Few approaches focus on complete business intelligence (BI) workflows and position the agents or agentic frameworks as platforms for data analysis \cite{weng2025datalabunifiedplatformllmpowered, weng2024insightlensaugmentingllmpowereddata, ma2023demonstrationinsightpilotllmempoweredautomated}  — combining query interfaces, tool libraries, and visualization modules. \cite{weng2025datalabunifiedplatformllmpowered, weng2024insightlensaugmentingllmpowereddata} are broader frameworks that include offline pre-processing stages to gather metadata for data understanding and schema mapping. \cite{hong2024datainterpreterllmagent, weng2025datalabunifiedplatformllmpowered, weng2024insightlensaugmentingllmpowereddata} focus on having modules that are specific to stages of insight generation such as SQL generation, data cleaning, chart generation, etc. These platform-centric agents prioritize user workflows — handling tasks like prompt interfaces, chart rendering, and multi-modal output —while treating reasoning as a black-box module abstracted behind orchestration layers. Even in works focusing on insight generation \cite{weng2024insightlensaugmentingllmpowereddata, sahu2025insightbenchevaluatingbusinessanalytics, ma2023demonstrationinsightpilotllmempoweredautomated}, the reasoning process is treated as a sequence of Q\&As on the data. While this is strong in guiding exploration, they lack explicit structured planning and execute using flat reasoning paths.

In particular, existing methods fall short in key areas that our work aims to address: (1) insufficient data exploration during early planning, (2) failure to detect procedural constraints as per the business rules (in the vastness of the context) \cite{shi2023largelanguagemodelseasily}, and (3) misalignment between planning and execution.

\section{Approach}
\label{sec:approach}
\begin{figure}[h]
    \centering
    \includegraphics[width=1\linewidth]{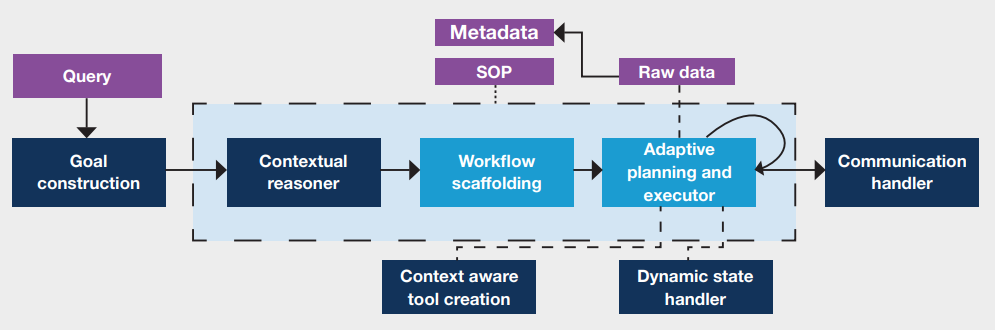}
    \caption{The workflow of sub-tasks for I2I-STRADA. A user query is first translated into a contextualized goal through a structured goal construction phase. This involves understanding the core analytical intent, identifying key entities and constraints, and outlining a preliminary solution approach—derived solely from the query. The goal is then refined and grounded using metadata and Standard Operating Procedures (SOPs) to ensure alignment with available data structures and domain-specific norms. Once contextualized, the goal enters a two-stage planning process. The workflow scaffolding module defines a high-level strategy, which guides the adaptive planner and executor—an iterative component that refines actions based on live data interaction. This core reasoning loop is supported by modules for dynamic tool creation and execution state management, while a communication handler delivers the final, user-aligned output in the required format. Refer to Algorithm \ref{alg:strada}.}
    \label{fig:workflow}
\end{figure}

\begin{algorithm}[ht]
\caption{I2I-STRADA: Structured Reasoning Agent for Data Analysis}
\label{alg:strada}
\begin{algorithmic}[1]
\Require User query $Q$, Raw data sources $D$, SOPs $S$, Instructions for handling data sources $I$
\Ensure Result for the user query in natural language $R$
\Statex
\Statex \textbf{I.} \textit{\textbf{(Offline step)}}: \textit{Prepare metadata from $D$, $S$ to support structured reasoning}
\Statex
\State $M \gets$ \Call{CreateMetadata}{$D$, $S$}
\Statex
\Statex \textbf{Main Procedure: \textsc{I2I-STRADA($Q, M, S, D, I$)}}
\Statex
    \Statex \textbf{II.} \textbf{Goal Construction}
    \State Analyze $Q$ and build belief state $B_0$ using question understanding, entities, constraints and solution approach
    \Statex
    \Statex \textbf{III.} \textbf{Contextual Grounding}
    \State Use metadata $M$ and SOPs $S$ to update belief $B_0 \rightarrow B$
    \Statex
    \Statex \textbf{IV.} \textbf{Workflow Scaffolding}
    \State Generate high-level plan $P = \{t_1, t_2, \dots, t_n\}$ based on $B$
    \State Initialize execution context $C_0$
    \Statex
    \Statex \textbf{V.} \textbf{Adaptive Planning and Execution}
    \Repeat
        \State $i \gets 1$
        \State Derive tool/code using $I$, $M$ and $C_{i-1} \rightarrow T_i(D)$
        \State Execute $T_i(D)$, observe results $r_i$
        \State Update execution context $C_i$
        \Statex Based on $C_i$:
        \If{$t_i$ complete}
            \State $i \gets i+1$
        \Else
            \State continue
        \EndIf
    \Until{$i$ = $n+1$}
    \Statex
    \Statex \textbf{VI. Results}
    \State Based on $C_n$, contextualize the results to the user query $Q$ and generate response $R$

    \State \Return $R$
\end{algorithmic}
\end{algorithm}

Our design is grounded in two key tenets: (1) progressive abstraction, where we preserve critical information while filtering noise at each stage; (2) multi step refinement, using a two-stage planning process to iteratively improve reasoning quality. This structured and modular approach enables robust and interpretable agent behavior in complex analytical settings.

In this section, we present the architecture and workflow of I2I-STRADA, detailing how each component contributes to a structured reasoning pipeline (see Figure \ref{fig:workflow}).

\textbf{Goal construction:} The initial step involves inferring the user's analytical goal directly from the given query. The agent constructs its "beliefs" about the data by extracting information solely from the query itself. This early identification of the problem type is essential for guiding subsequent data exploration, and building belief from scratch ensures that the agent considers every detail relevant to the request. The outcome of this step consists of:
\begin{itemize}
    \item Question understanding - Understand the core intent of the user
    \item Entity extraction - identifying relevant data points, dimensions, or concepts mentioned in the query;
    \item Generic solution approach - outlining a preliminary high-level strategy; and
    \item Constraints - detailing any specific limitations or conditions provided.
\end{itemize}
Refer to Appendix \ref{sec:AppendixA} for the prompt.

\textbf{Contextual reasoner:} Acting as a bridge between the initial understanding and a plan of action, the module grounds the analysis using contextual information. It references metadata of the data systems and applicable SOPs to refine the solution approach derived from the inferred goal and constructed belief. Utilizing these inputs helps ensure the resulting plan is not only aligned with the user's request but also key procedural requirements and constraints. Refer to Appendix \ref{sec:AppendixB} for the prompt.

\textbf{Two Planning stages:}
    \begin{itemize}
        \item \textbf{Workflow scaffolding:} The Workflow Scaffolding is the generator of a global plan of action. This plan is formulated before the agent interacts with the actual data. This high-level plan serves as the foundational workflow or 'scaffold' that guides the adaptive executor, allowing for dynamic execution while ensuring the analysis adheres to the defined overall problem-solving approach. Refer to Appendix \ref{sec:AppendixC} for the prompt.
        \item \textbf{Adaptive planning and executor:} It is an iterative module that generates execution-level plans aligned with the scaffolded workflow. It dynamically adjusts subsequent steps based on prior execution results, including actual data exploration and intermediate outcomes. This adaptability is necessary as complex tasks require data interaction to inform planning. The adaptive planner ensures alignment with the scaffold and tracks plan status iteratively. The execution involves writing code snippets in Python and executing them in a sandbox. The context of the execution carries through all the iterations. Refer to Appendix \ref{sec:AppendixD}, \ref{sec:AppendixE} for the prompts.
    \end{itemize}

\textbf{Context aware tool creation:} The module utilizes metadata (types of data sources involved) and instructions (how to process the data, recommended libraries to use etc.) to dynamically create data processing tools and scripts on the fly. This is key to analyzing heterogeneous data sources effectively and extends the solution's applicability to Bring Your Own (BYO) data sources.

\textbf{Dynamic State Handler:} Acts as the agent's dynamic working memory, essential due to adaptive execution planning. It maintains the execution context across iterations (includes updating variables) and provides runtime debugging capabilities.

\textbf{Communication Handler:} Manages the presentation of results, ensuring they address user goals and conform to required formatting. It converts raw output based on guidelines or query context, making information clear and relevant.

\section{Evaluation}
\label{sec:eval}
We evaluate our solution on two recent benchmark datasets to validate the generalizability of the approach. The closest
benchmark that aligned with the idea of procedural knowledge driven multi-source data analysis was DABstep \cite{egg2025dabstepdataagentbenchmark}. The
second benchmark dataset is DABench \cite{hu2024infiagentdabenchevaluatingagentsdata}. This dataset has a stronger focus on statistics and data science. These two
datasets provide a wide spectrum of concepts to test the efficacy of agentic approach for data analysis.

\subsection{Results on DABstep benchmark}
The DABstep dataset\cite{egg2025dabstepdataagentbenchmark}, developed by Adyen in collaboration with Hugging Face contains tasks that test reasoning over
financial and operational data. It comprises over 450 tasks that simulate real-world analytical workflows common in
financial services, such as interpreting transaction records, navigating policy documentation, and reconciling structured
and unstructured data sources.

We used Anthropic’s Claude 3.5 Sonnet in our agentic workflow. Our agent outperforms several SOTA data science agents as well as baselines built using ReACT \cite{yao2023reactsynergizingreasoningacting} framework with an accuracy of 80.56\% on easy tasks and 28.04\% on hard tasks. Refer to table \ref{tab:dabstebleaderboard}.

\begin{table}[h]
    
    \centering
    \resizebox{\textwidth}{!}{ 
    \begin{tabular}{lcclll}

    \toprule
    \textbf{Agent} & \textbf{Easy Level Accuracy} & \textbf{Hard Level Accuracy} & \textbf{Organization} & \textbf{Model Family} & \textbf{Date} \\ 
    \midrule
    \textbf{Mphasis-I2I-Agents} & \textbf{80.56\%} & \textbf{28.04\%} & \textbf{\textit{(Ours) Mphasis Limited}} & \textbf{claude-3-5-sonnet} & \textbf{10/4/2025} \\ 
    DICE & 75.00\% & 27.25\% & Microsoft & o3-mini & 17-04-2025 \\ 
    O4-mini Reasoning Prompt Baseline & 76.39\% & 14.55\% & Hugging Face & OpenAI o4-mini & 22-04-2025 \\
    Claude 3.7 Sonnet ReACT Baseline & 75.00\% & 13.76\% & Hugging Face & claude-3-7-sonnet & 7/4/2025 \\ 
    Gemini Data Science Agent & 61.11\% & 9.79\% & Google & Gemini 2.0 Flash & 10/2025 \\ 
    Claude 3.5 Sonnet ReACT Baseline & 77.78\% & 9.26\% & Adyen & claude-3-5-sonnet & 23-01-2025 \\ 
    Deepseek V3 ReACT Baseline & 66.67\% & 5.56\% & Adyen & Deepseek v3 & 23-01-2025 \\ 
    Llama 3.3 70B ReACT Baseline & 68.06\% & 3.70\% & Adyen & Llama 3.3 70B Instruct & 23-01-2025 \\ 
    \bottomrule
    \end{tabular}
    }
    \caption{DABstep leaderboard. View the live leaderboard at: \url{https://huggingface.co/spaces/adyen/DABstep}}
    \label{tab:dabstebleaderboard}
\end{table}

Where our agent succeeds:  
\begin{itemize}
    \item Improved planning and failure handling when writing code 
    \item Sensitive to rules mentioned in the SOP 
    \item Planning without overthinking (Easy tasks require simple plans)
\end{itemize}

Where we see chances to improve: 
\begin{itemize}
\item The agent seems inconsistent when applying SOP rule related to handling of “Null” values. It correctly interprets empty lists (i.e [] ) as “Null” always but on several occasions, when a field is explicitly “null”/”None”, it fails to apply this rule. This seems to be an interpretation problem with Claude 3.5 Sonnet as it focuses attention on a single example given in the SOP.  
\end{itemize}
Appendix \ref{sec:AppendixF} presents our agent’s trace on one hard task. The example represents the attention to detail arising out of multi-stage refined planning. The rest of the reasoning traces are available on Hugginface DABstep submissions for reference. 

\subsection{Results on DABench benchmark}
The InfiAgent-DABench benchmark\cite{hu2024infiagentdabenchevaluatingagentsdata}, is specifically designed to evaluate large language model (LLM)-based agents on end-to-end data science tasks across a variety of real-world domains (Marketing, Finance, Energy etc.). The core of the benchmark is the DAEval dataset, comprising 257 open-ended data analysis questions associated with 52 diverse CSV files collected from public sources.  
The concepts covered by the tasks include - Summary Statistics, Feature Engineering, Correlation Analysis, Machine Learning, Distribution Analysis, Outlier Detection and Comprehensive Data Preprocessing. The dataset doesn’t have SOPs. We hence provided just the definitions of the tasks given by  as SOP input.

\begin{table}[h]
    \centering
    \begin{tabular}{lcl}
    \hline
    \toprule
    \textbf{Agent} & \textbf{Accuracy} & \textbf{Model Family} \\ 
    \midrule
    Data Interpreter \cite{hong2024datainterpreterllmagent} & 94.93\% & GPT-4o \\
    \textbf{Mphasis-I2I-Agents (Ours)} & \textbf{90.27\%} & \textbf{claude-3-5-sonnet} \\ 
    Datawise Agent \cite{you2025datawiseagentnotebookcentricllmagent} & 85.99\% & GPT-4o \\ 
    Data Interpreter \cite{hong2024datainterpreterllmagent} & 73.55\% & GPT-4 \\ 
    AgentPoirot \cite{sahu2025insightbenchevaluatingbusinessanalytics} & 75.88\% & GPT-4 \\ 
    DataLab \cite{weng2025datalabunifiedplatformllmpowered} & 75.10\% & GPT-4 \\ 
    \bottomrule
    \end{tabular}
    
    \caption{Performance comparison on DABench}
    \label{tab:dabench_leaderboard}
\end{table}

The accuracy metric shown in table \ref{tab:dabench_leaderboard} is accuracy by question (ABQ).  The numbers are as reported in the respective papers, and we haven’t attempted to replicate them.  Additionally, we have picked only the best results from these papers to compare against. 

Where our agent succeeds: 
\begin{itemize}
    \item Single/Multi source, the same workflow without any modifications produces consistently SOTA results. 
    \item The exact nature of the data analysis task doesn’t affect the performance. (Domain specific or pure statistical/data science based) 
\end{itemize}

Where we see chances to improve: 
\begin{itemize}
    \item When applying machine learning algorithms, the choice of hyperparameters often results in different results. This could be corrected by providing an appropriate procedure document.  
\end{itemize}
Appendix \ref{sec:AppendixG} presents our agent’s traces on a hard task. 

\section{Conclusion}
In this work, we have presented an agentic system design to address the multifaceted challenges of data analysis in
real-world scenarios. Our approach leverages a structured workflow composed of specialized sub-tasks, each dedicated
to a distinct aspect of reasoning and planning. The multi-step context refinement process, supported by contextual tool creation
ensures that the agent can handle heterogeneous data sources, perform complex intermediate calculations, and support a
wide array of analytical queries. 

Our evaluation on the DABstep and DABench benchmarks demonstrates the effectiveness and generalizability of our
agent. On DABstep, our agent outperforms other SOTA solutions, particularly excelling in planning and failure handling
when writing code and adhering to SOPs. On DABench, our agent shows robustness across diverse domains and data
analysis tasks, maintaining high accuracy without modifications to its workflow. Additionally, our approach substantially addresses the reasoning limitations of LLMs in complex analytical scenarios \cite{illusion-of-thinking}.

In conclusion, we believe that this approach can further the development of fine-tuned reasoning models to be used in
agentic systems capable of performing comprehensive data analysis.

\bibliographystyle{unsrt}  
\bibliography{references}  

\appendix
\section{Appendix - Prompt for Goal construction}
\label{sec:AppendixA}
\begin{verbatim}
You are given a user query. You have to extract the following things from the query 
and context provided to you:
    Question understanding: What do you understand from the question.
    Entity extraction: Key entities in the question.
    Solution approach: How to solve the question in general
    Constraints: If any constraints or any additional details which are given in the context 
which you have to take care while answering the questions
\end{verbatim}

\section{Appendix - Prompt for Contextual reasoner}
\label{sec:AppendixB}
\begin{verbatim}
Relevant chunks from context: Extract relevant chunks(exact match) from 
the context which help you get the answer
    The context is given by:
    <context>
    {content}
    {content2}
    </context>
The user query is given by:
    <user query>
    {query}
    </user query>
The current understanding/belief is given by:
    <belief>
    {belief}
    </belief>

Provide a solution approach: How to solve the problem using the context given to you
\end{verbatim}

\section{Appendix - Prompt for Workflow scaffolding}
\label{sec:AppendixC}
\begin{verbatim}
You are a chatbot who has to create a checklist for a downstream 'plan executor' pipeline. 
You have to create checklist to solve user queries based on the information 
available in the context and the metadata given to you.

The context is given by:

<context>
{context_for_planner}
</context>

The metadata is given by:

<metadata>
{metadata}
</metadata>

These are the sources of the data which you have:
{files_list}

Follow these instructions:
<instructions>
{custom_instructions}
</instructions>

The output should be only a parsable JSON whose format is given by:

<output_format>
{output_format}
</output_format>

The user query is:
<user_query>
{query}
</user_query>      
\end{verbatim}

\section{Appendix - Prompt for Adaptive planning and executor (1) }
\label{sec:AppendixD}
\begin{verbatim}
You are an AI assistant. You have to execute a plan given 
by a upstream 'planner agent' for completing a user query.
The plan you have to follow is given by:

<plan>
{plan}
</plan>

The files given to you are:
{files_list}

This is the metadata for the files which given to you. 
When making a query make sure the column names, values 
and everything follows the metadata.

<metadata>
{metadata}
</metadata>

Follow these instructions:
<instructions>
{instructions}
</instructions>

The output should be only a parsable JSON whose format is given by:
<output_format>
{output_format}
</output_format>
\end{verbatim}

\section{Appendix - Prompt for Adaptive planning and executor (2) }
\label{sec:AppendixE}
\begin{verbatim}
The response for the query is given by:
{response}

Instructions:
1. Reflect on the output and take the next step according to the response as well as the plan.
2. If you get a error on your code, rewrite the code and make sure you have no undefined variables.
3. In case of writing code, every query is independent. \
So make sure you define variable names before using them directly as variables only have local scope.
4. You will put the plan status as "completed" when you have the final answer with yourself and you have to give it to user. 
5. The response has to strictly follow the output format given to you.
\end{verbatim}

\section{Appendix - Example trace of I2I-STRADA on DABstep}
\label{sec:AppendixF}
\textbf{Hard task – Task ID: 1434}

\textbf{Question:} What is the most expensive MCC for a transaction of 5 Euros, in general? If there are many MCCs with the same value, list all of them. Provide a list as an output even if it is one element.

\textbf{Guideline:} Answer must be a list of values in comma-separated list, eg: A, B, C. If the answer is an empty list, reply with an empty string. If a question does not have a relevant or applicable answer for the task, please respond with ‘Not Applicable’.

\textbf{Trace:}
\begin{figure}
    \centering
    \includegraphics[width=1\linewidth]{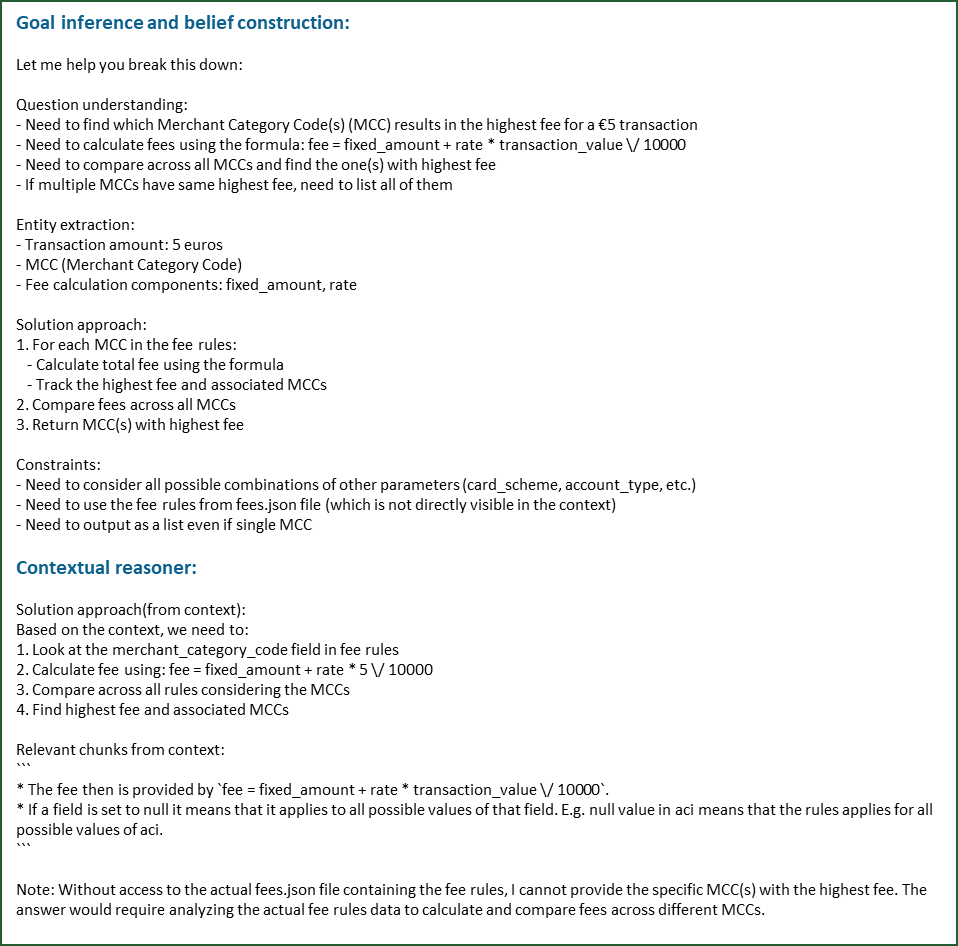}
\end{figure}
\begin{figure}
    \centering
    \includegraphics[width=1\linewidth]{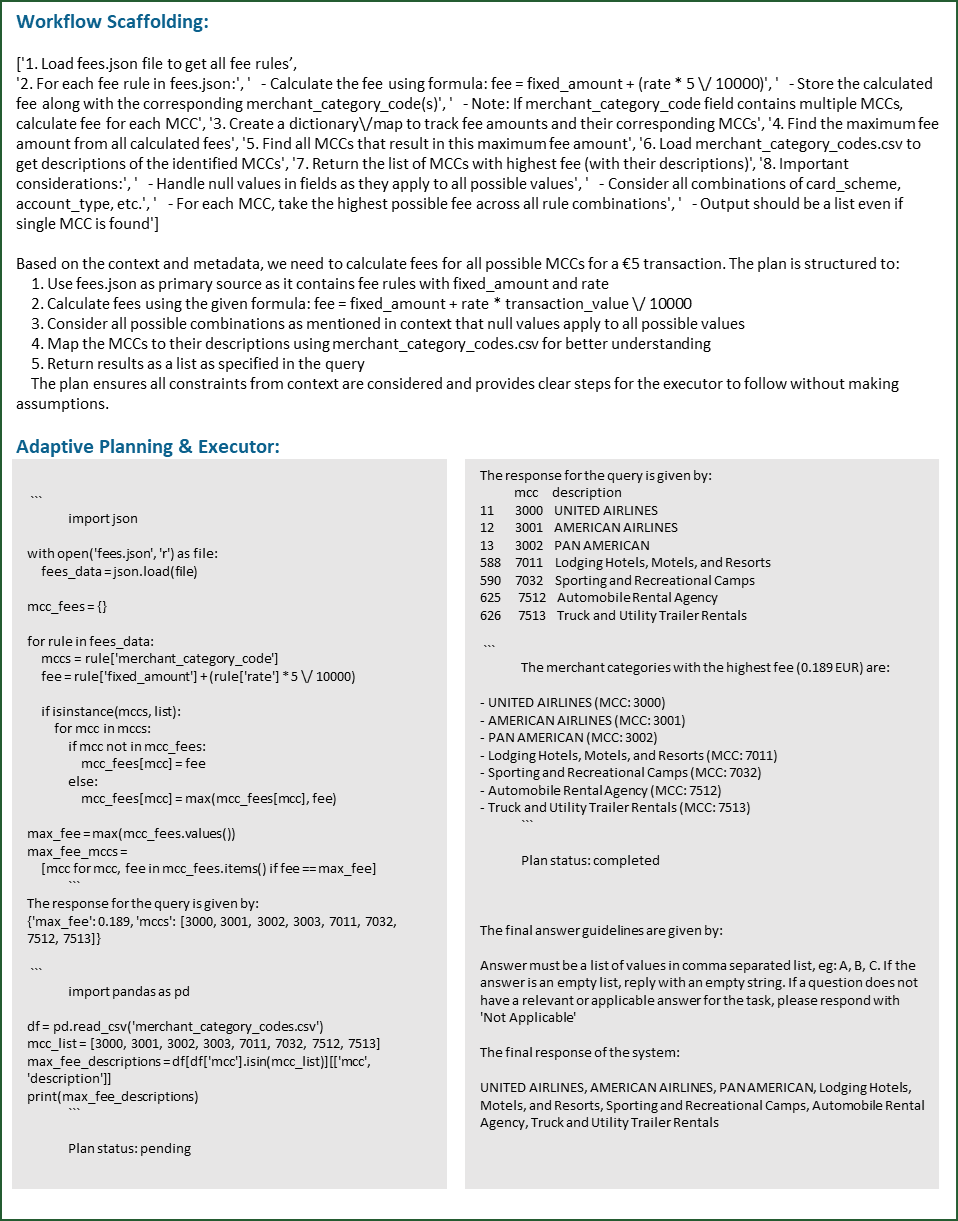}
    \caption{Trace of the agent for task 1434 from DABstep dataset}
    \label{fig:tracedabstep}
\end{figure}

\section{Appendix - Example trace of I2I-STRADA on DABench}
\label{sec:AppendixG}
\textbf{Hard task – Task ID - 28}

\textbf{Question:} Perform comprehensive data preprocessing on the dataset, including cleaning, transformation, and handling of missing values. Handle the missing values in the 'age', 'sex', and 'region' columns by removing the corresponding rows. Transform the 'sex' and 'smoker' columns to binary format (0 and 1). Normalize 'age', 'bmi', 'children', and 'charges' columns. Report the mean of each column after the preprocessing.

\textbf{Trace:}
\begin{figure}
    \centering
    \includegraphics[width=1\linewidth]{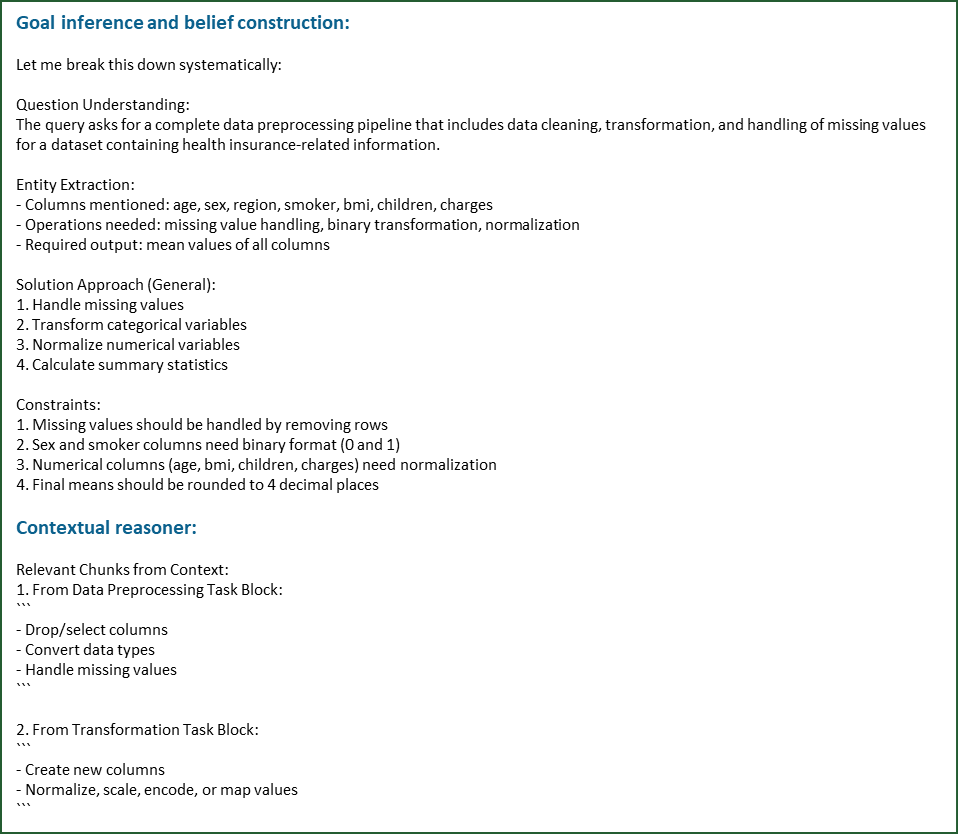}
\end{figure}
\begin{figure}
    \centering
    \includegraphics[width=1\linewidth]{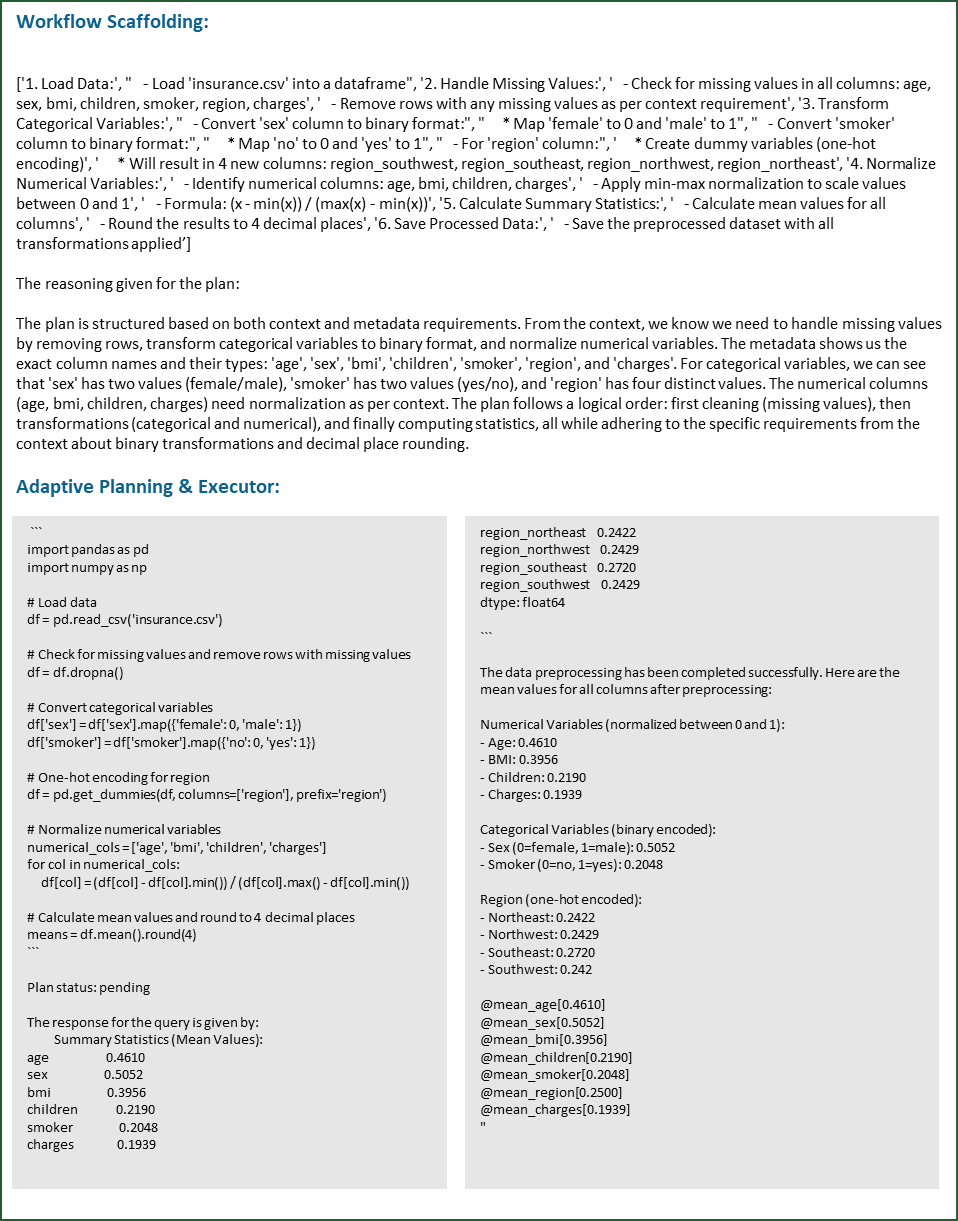}
    \caption{Trace of the agent for task 28 from DABench dataset}
    \label{fig:tracedabench}
\end{figure}
\end{document}